\documentclass[lettersize,journal]{IEEEtran}
\usepackage{amsmath,amsfonts}
\usepackage{algorithmic}
\usepackage{algorithm}
\usepackage{array}

\usepackage{textcomp}
\usepackage{stfloats}
\usepackage{url}
\usepackage{verbatim}
\usepackage{graphicx}
\usepackage{cite}
\usepackage{bbding}
\usepackage{booktabs}
\usepackage{color}

\usepackage{adjustbox}
\usepackage{booktabs}
\usepackage{algorithm}
\usepackage{algorithmic}
\usepackage{multirow}
\usepackage{multicol}
\usepackage{dsfont}
\usepackage{amssymb}
\usepackage{float}
\usepackage{hyperref}
\usepackage[table,xcdraw]{xcolor}

\begin{document}
{
\title{NODE-Adapter: Neural Ordinary Differential Equations for Better Vision-Language Reasoning}

\author{Yi~Zhang,~Chun-Wun~Cheng,~Ke~Yu,~Zhihai~He,~Carola-Bibiane Schönlieb, and~Angelica~I.~Aviles-Rivero
\thanks{Yi Zhang is with Harbin Institute of Technology, Harbin, China (e-mail: zhangyi2021@mail.sustech.edu.cn).}
\thanks{Yi Zhang, Ke Yu, and Zhihai He are with the Department of Electrical and Electronic Engineering, Southern University of Science and Technology, Shenzhen, China (e-mail: yuk2020@mail.sustech.edu.cn; hezh@sustech.edu.cn).}
\thanks{Chun-Wun Cheng, Carola-Bibiane Schönlieb and Angelica I. Aviles-Rivero are with the Department of Applied Mathematics and Theoretical Physics, University of Cambridge, Cambridge, UK (e-mail: cwc56@cam.ac.uk; cbs31@cam.ac.uk; ai323@cam.ac.uk).}
\thanks{Corresponding author: Angelica I. Aviles-Rivero (ai323@cam.ac.uk).}

}



\maketitle
\begin{abstract}
In this paper, we consider the problem of prototype-based vision-language reasoning problem. We observe that existing methods encounter three major challenges: 1) escalating resource demands and prolonging training times, 2) contending with excessive learnable parameters, and 3) fine-tuning based only on a single modality. These challenges will hinder their capability to adapt Vision-Language Models (VLMs) to downstream tasks.
Motivated by this critical observation, we propose a novel method called \textbf{NODE-Adapter}, which utilizes \textbf{N}eural \textbf{O}rdinary \textbf{D}ifferential \textbf{E}quations for better vision-language reasoning. To fully leverage both visual and textual modalities and estimate class prototypes more effectively and accurately, we divide our method into two stages: cross-modal prototype construction and cross-modal prototype optimization using neural ordinary differential equations. Specifically, we exploit VLM to encode hand-crafted prompts into textual features and few-shot support images into visual features. Then, we estimate the textual prototype and visual prototype by averaging the textual features and visual features, respectively, and adaptively combine the textual prototype and visual prototype to construct the cross-modal prototype. To alleviate the prototype bias, we then model the prototype optimization process as an initial value problem with Neural ODEs to estimate the continuous gradient flow. 
Our extensive experimental results, which cover few-shot classification, domain generalization, and visual reasoning on human-object interaction, demonstrate that the proposed method significantly outperforms existing state-of-the-art approaches.
\end{abstract}


\section{Introduction}
\label{sec:intro}

\begin{figure}[t]
 \setlength{\abovecaptionskip}{0.2cm}
 \setlength{\belowcaptionskip}{-0.1cm}
\centerline{\includegraphics[width=\linewidth]{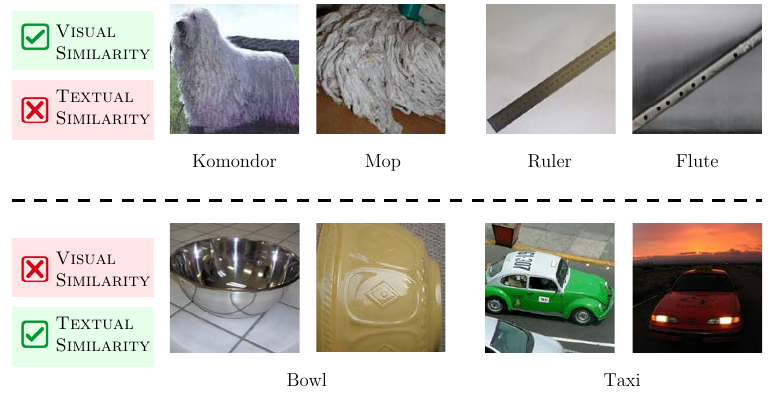}}
\caption{Classes exhibit distinct visual and textual feature spaces respectively. Images from different classes may share similar visual features but differ in textual features, and conversely, images from the same class may showcase diverse visual features. Our goal is to leverage both modalities to enhance performance in few-shot classification scenarios.}
\label{fig:intro}
\vspace{-10pt}
\end{figure}

\IEEEPARstart{R}{ecent} research on pre-trained Vision-Language Models (VLMs), like CLIP \cite{radford2021learning}, has shown a promising path for foundational models to excel in numerous open-vocabulary tasks. These models leverage their understanding of visual concepts from extensive image-text pairs, showcasing impressive abilities across various downstream tasks, often with zero or few-shot learning \cite{radford2021learning}.

While the zero-shot CLIP model exhibits strong performance across diverse visual tasks, its pre-trained nature limits its ability to adapt to new, unseen domains. Consequently, many studies aim to enhance these pre-trained VLMs for specific downstream tasks by developing learnable prompts based on training examples. These fine-tuning methods can be categorized into input-stage prompting methods\cite{zhou2022learningCoOp,lu2022promptProDA,zhou2022conditional,chuang2023debiasingDebiasprompt,shu2022testTPT,huang2022unsupervisedUPL,loedeman2022promptPGN,chenplotPLOT}, and feature-stage fine-tuning methods\cite{gao2021clipadapter,pantazis2022svlSVL-Adapter,sung2022vlVL-Adapter,zhu2023notAPE,zhang2022tipTip-Adapter,zhang2023promptCaFo}.

Specifically, input-stage prompting modifies the textual classifier for downstream tasks by incorporating trainable prompts at the input level, significantly outperforming zero-shot CLIP when using few-shot examples, as shown in studies like CoOp~\cite{zhou2022learningCoOp} and CoCoOp~\cite{zhou2022conditional}. However, this approach has a notable drawback in Vision-Language Models (VLMs): it requires data to be processed through the textual encoder in each training iteration, increasing the demand for resources and extending training times.

Conversely, utilizing the textual encoder just once, the feature stage typically fine-tuning improves the textual classifiers or visual features through straightforward yet effective feature modulation for particular tasks at the output stage.
For example, CLIP-Adapter\cite{gao2021clipadapter} employs a single bottleneck layer to modify both textual and visual embeddings in VLMs, achieving a 3.02\% improvement over zero-shot CLIP in the one-shot ImageNet setting. Similarly, TaskRes\cite{yu2022taskTaskRes} uses learnable, task-specific parameters that serve as modality-independent residuals to tweak the textual embeddings. Another emerging approach\cite{zhang2022tipTip-Adapter,zhang2023promptCaFo,zhu2023notAPE} involves enhancing the existing knowledge for downstream tasks by integrating CLIP with other pre-trained major vision or language models, such as DINO\cite{caron2021emergingDINO}, and GPT\cite{brown2020languageGPT-3}. However, there are two limitations in most feature-stage fine-tuning works on VLMs' adaptation: 1) Focusing solely on acquiring task-specific knowledge from a single modality, such as TaskRes~\cite{yu2022taskTaskRes} and Tip-Adapter~\cite{zhang2022tipTip-Adapter}, where fine-tuning is based solely on visual or textual features. However, as shown in Figure \ref{fig:intro}, we observe that images from different classes may share similar visual features but differ in textual features, and conversely, images from the same class may share similar textual features but exhibit diverse visual features. Thus, fine-tuning based only on a single modality will hinder its capability to adapt to downstream tasks.
2) Suffering from excessive learnable parameters, such as Tip-Adapter\cite{zhang2022tip} and Tip-X\cite{udandarao2022sus-x}, these approaches incorporate prior knowledge directly into the training process. However, they are often cumbersome because they require a large cache size and an extensive number of learnable parameters.

\begin{figure}[t]
\begin{center}
\centerline{\includegraphics[width=\linewidth]{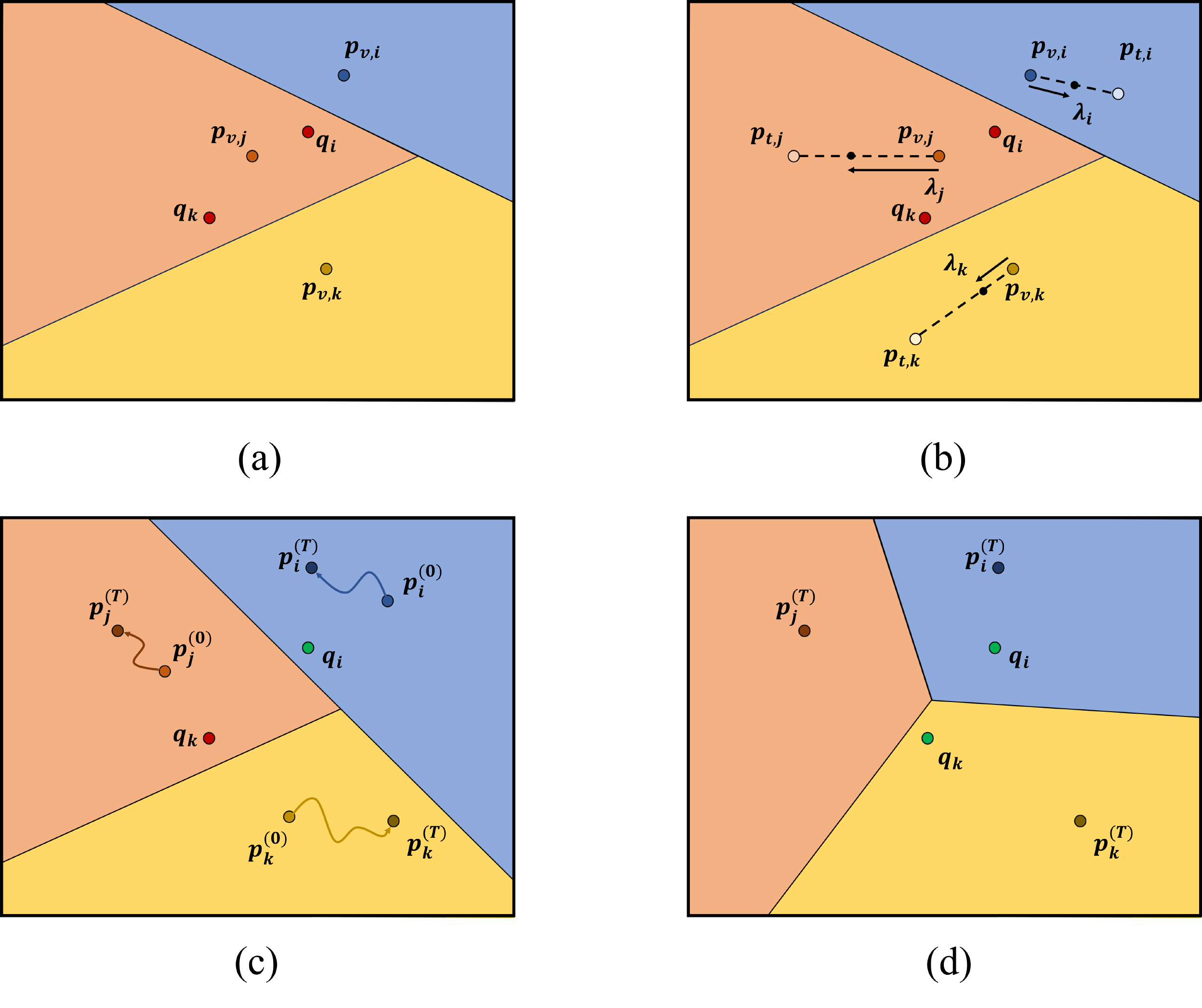}}
\caption{Illustration of prototype rectification. $\mathbf{q}_i$ and $\mathbf{q}_k$ are query samples of class $i, k$, respectively. For class i, j and k,  $\{\mathbf{p}_{v,i},\mathbf{p}_{v,j},\mathbf{p}_{v,k}\}$ are the visual prototypes,  $\{\mathbf{p}_{t,i},\mathbf{p}_{t,j},\mathbf{p}_{t,k}\}$ are textual prototypes and $\{\mathbf{p}_i,\mathbf{p}_j,\mathbf{p}_k\}$ are the cross-modal prototypes. 
\textbf{(a)} Initially the two query samples are misclassified. \textbf{(b)} Cross-modal prototypes corrects the classification of $\mathbf{q}_i$. \textbf{(c)} The cross-modal prototypes are further rectified by Neural ODE. Hence, in \textbf{(d)}, both of the query samples are corrected at time $T$.
}
\label{fig:rect}
\end{center}
\vspace{-20pt}
\end{figure}

To mitigate the above limitations, in this paper, we propose a novel method called \textbf{NODE-Adapter} which utilizes \textbf{N}eural \textbf{O}rdinary \textbf{D}ifferential \textbf{E}quations for better vision-language reasoning. We optimize the estimated class prototype with two stages: cross-modal prototype construction and cross-modal prototype optimization with Neural Ordinary Differential Equations. As shown in figure \ref{fig:rect}, pictures (a) and (b) shows the process of cross-modal prototype construction. Picture (c) represents the process of cross-modal prototype optimization with Neural ODEs and picture (d) displays the final results. Specifically, we exploit VLM to encode hand-crafted prompts into textual features and few-shot support images into visual features. We estimate the textual prototype and visual prototype by averaging the textual features and visual features, respectively, and adaptively combine the textual prototype and visual prototype to construct the cross-modal prototype. However, this approach faces a prototype bias issue due to the discrpancy between the calculated mean and the actual prototypes. This bias stems from the insufficient number of labeled samples, which hinders accurate mean estimation for the prototypes.

To address prototype bias, we propose a novel Neural ODEs model for optimizing prototypes as continuous-time dynamics. Our approach is motivated by recognizing that gradient flow in gradient descent can be analogized to the Euler method, a numerical approach for solving Ordinary Differential Equations (ODEs). This understanding allows us to model continuous gradient flow using an ODEs framework and apply Neural ODEs to refine prototypes more effectively for vision-language tasks. Specifically, we develop a Neural ODEs module to estimate continuous-time gradient flow dynamics for prototype optimization. Starting from an initial estimated cross-modal prototype, our method iteratively refines this prototype by solving the Neural ODEs to achieve an optimized cross-modal prototype.
Neural ODEs utilize a black box ODE solver, enabling the application of the adjoint-sensitive method. As a result, this approach ensures constant memory cost, thereby enhancing computational efficiency.
After obtaining the optimized cross-modal prototypes, the class membership of the test image is determined by the nearest neighbor strategy. 
Our contributions can be summarized as below:
\begin{itemize}
    \item We propose a state-of-the-art method called NODE-Adapter, which provides a novel prototype-based approach to adapt CLIP for downstream tasks.
    \item We first adaptively synthesize textual features and visual features to construct a cross-modal prototype, then exploit Neural Ordinary Differential Equations to alleviate the prototype bias.
    \item We conduct extensive experiments on few-shot classification, domain generalization, and visual reasoning for human-object interaction. Our results showed that our proposed method significantly outperforms existing state-of-the-art approaches.
    
\end{itemize}

The structure of the remaining part is organized as follows: Section \ref{sec:related_work} reviews the related works. In Section \ref{sec:method}, we present the review of CLIP and the proposed NODE-Adapter method. Details of the experimental setup and an analysis of the experimental results are provided in Section \ref{sec:experiments}. Lastly, Section \ref{sec:conclusion} concludes the paper and suggests directions for future work.}

\section{Related Work}
In this section, we review related works on vision-language models, adaptation of VLMs, Human Object Interaction (HOI) detection and neural ordinary differential equations.
\label{sec:related_work} 
\subsection{Vision-Language Models}
Extensive efforts have been devoted to the development of large-scale pre-trained Vision-Language Models (VLMs), aiming to acquire comprehensive visual representations guided by natural language supervision \cite{lei2015predicting,gomez2017self,sariyildiz2020learning,desai2021virtex,radford2021learning}. Current research endeavors have delved into investigating the semantic alignment between linguistic and visual modalities, leveraging abundant image-text pairs available online \cite{jia2021scaling,radford2021learning,yu2022coca}. Noteworthy examples include CLIP \cite{radford2021learning}, derived through contrastive learning on 400 million meticulously curated image-text pairs, and ALIGN \cite{jia2021scaling}, which exploits 1.8 billion noisy image-text pairs sourced from raw alt-text data. Numerous other initiatives exploring the realm of large-scale VLMs include CoCa \cite{yu2022coca}, SimVLM \cite{wang2022simvlm}, Florence \cite{yuan2021florence}, BEiT \cite{wang2022image}, Flamingo \cite{alayrac2022flamingo}, PaLI \cite{chen2022pali}.
Demonstrations by researchers underscore the versatility of large-scale pre-trained VLMs in tackling various cross-modal alignment, zero-shot, and few-shot image recognition tasks \cite{radford2021learning,zhang2021vt,zhou2022learning}. 

\subsection{Adaptation of VLMs for Few-shot Learning}
Adaptation is essential for applying Vision-Language Models (VLMs) to diverse downstream tasks. This study specifically targets few-shot image classification, categorizing recent works into prompt learning and adapter-style methods.

\textbf{Prompt learning methods.} They draw inspiration from the success of prefix-tuning in natural language processing \cite{deng2022rlprompt,jiang2020can}. Pioneering this domain, CoOp \cite{zhou2022learning} augments prompt context by optimizing it through trainable vectors. Extending this approach, Zhou \textit{et al.} \cite{zhou2022conditional} addresses generalization issues for unseen classes by conditioning vector generation on each image. To prevent prompt learning from forgetting general knowledge and ProGrad \cite{zhu2022prompt} suggests updating prompts with well-aligned gradients.  Other studies explore prompt learning for VLMs; for instance, CPL \cite{zhang2024concept} achieved improved consistency among visual and text modalities, while ProDA \cite{lu2022prompt} captures diverse prompt distributions from a limited support set, accommodating varied visual representations. DPT \cite{xing2023dual} suggests simultaneously learns text and visual prompts.

\textbf{Adapter-style methods.} This family of methods is influenced by parameter-efficient fine-tuning techniques \cite{zhang2020side}, directly modify the representations generated by CLIP's visual and text encoders. For instance, CLIP-Adapter \cite{gao2021clip} introduces an additional feature adapter to enhance conventional fine-tuning outcomes. TT-DNA-Adapter performs throughout the testing period \cite{zhang2024test}. Tip-Adapter \cite{zhang2022tip} enhances results by creating a key-value cache model with low-shot samples.
APE \cite{zhu2023not} effectively adapts CLIP for few-shot classification by fine-tuning its pre-trained knowledge in visual representations. GraphAdapter \cite{li2024graphadapter} enhances the textual adapter by explicitly incorporating the dual-modality structure knowledge using a dual knowledge graph. and CaFo \cite{zhang2023prompt} incorporates diverse prior knowledge from various pretraining paradigms. Our proposed approach falls within the adapter-style methods category. 

\subsection{Human Object Interaction Detection.}
Human-Object Interaction (HOI) detection is a fundamental
task in computer vision, focusing on identifying interactions between humans and objects within an image \cite{gkioxari2018detecting,wang2020learning}.  In recent years, various approaches have been proposed to address this task. InteractNet \cite{gkioxari2018detecting} proposes a novel human-centric approach to detecting (human, verb, object) triplets in everyday photos.
Chao \textit{et al.} \cite{chao2018learning} proposes a new benchmark dataset and a novel Human-Object Region-based Convolutional Neural Network (HO-RCNN) approach that leverages a novel DNN input called Interaction Pattern to improve the performance of HOI detection.
Wang \textit{et al.} \cite{wang2020learning} treats HOI detection as a keypoint detection and grouping problem, and proposes a novel fully-convolutional approach that directly detects the interactions between human-object pairs by predicting interaction points and associating them with human and object detections.
HOITrans \cite{zou2021end} directly predicts HOI instances in parallel by reasoning about the relations of objects and humans from global image contexts, thus eliminating the need for many hand-designed components.

\textbf{Visual Reasoning for Human Object Interaction.}
The ability to understand visual relationships is a crucial aspect of how humans perceive the visual world. In this context, Bongard-HOI \cite{jiang2022bongard} is a recently introduced benchmark emphasizing the compositional learning of HOIs from natural images, posing a substantial challenge to current visual recognition models. The leading detection model, HOITrans \cite{zou2021end}, achieves only 62\% accuracy in few-shot binary prediction tasks.
However, CLIP-based TPT \cite{shu2022tpt} and BDC-Adapter \cite{zhang2023bdc} has shown promising results without training on the training split, which indicates a new paradigm of multi-modal reasoning for HOI. We evaluate our proposed NODE-Adapter method on this task to demonstrate its effectiveness in visual relationship reasoning.

\subsection{Neural Ordinary Differential Equation}

Neural Ordinary Differential Equations (Neural ODEs) \cite{chen2018neural} represent an innovative form of deep implicit learning, conceptualized as a continuous extension of Residual Networks \cite{he2016deep}. In Neural ODEs, the evolution of the hidden state is continuous and governed by an Ordinary Differential Equation, modeled by a neural network:
$\frac{\mathrm{d}\mathbf{\textbf{h}}^{}}{\mathrm{d}t} = f_{\theta}(\textbf{h}(t), t)$, where $f_{\theta}$ is a neural network parametrized by $\theta$ . Specifically, $f_{\theta}$ is a standard deep neural network consisting of multiple layers, including linear (fully connected) or convolutional (CNN). 
The forward pass, in this context involves solving an ODE initial value problem, which can be efficiently solved by a black box ODE solver. Meanwhile, the computation of the gradient employs the adjoint sensitivity method, noted for its advantage of  constant $O(1)$ memory cost. Notably, this paradigm has demonstrated efficacy across diverse domains, including applications such as forecasting irregular time series \cite{rubanova2019latent}, medical images segmentation \cite{cheng2023continuous}, diffusion model \cite{ordonezmissing} and few shot learning \cite{zhang2022metanode}. Despite the broad success of these varied applications,  a significant gap remains in the
literature concerning the use of Neural ODEs for Cross-modal
FSL. In this work, we propose a novel approach that employs a Neural ODE-based deep learning framework to refine prototypes within the context of Cross-modal FSL. 
The major advantage of our proposed method lies in its ability to capture prototype dynamics continuously, allowing a higher performance due to the smaller step size, thereby enhancing the overall efficacy of Cross-modal Few-Shot Learning.

\section{Methodology}
\label{sec:method}
In this section, we present our proposed NODE-Adapter, Neural Ordinary Differential Equations for Better Visual-Language Reasoning in detail.
\subsection{Preliminaries}
\textbf{Revisiting  Contrastive Language-Image Pre-training (CLIP).}
CLIP comprises two parallel encoders: one for image processing, typically ResNet \cite{he2016deep} or ViT \cite{dosovitskiy2020image}, and another for text, based on transformer architectures. During training, a contrastive loss function is employed to encourage similarity between the image and text feature vectors, aligning both modalities in a joint embedding space. The CLIP model is represented as $\{E_t, E_v\}$, where $E_t$ is the text encoder and $E_v$ is the image encoder. Post-training, CLIP supports zero-shot application in downstream tasks using hand-crafted prompts. For image classification, given a test image $X_{test}$ belonging to a specific class $y$, where $X_{test}\in \mathbb{R} ^{C\times H\times W}$ and $y \in \mathbb{R} ^ N$ for an $N$-class problem, each class $y_i$ in the set $Y=\{y_1, y_2, \cdots, y_N\}$ is combined with a prompt $\rho =$ "\texttt{a photo of,}" to form class-specific textual inputs $\{\rho; y_i\}$. Text features $\{t_1, t_2, \cdots, t_N\}$ are extracted by $E_t$, where $t_i = E_t(\{\rho; y_i\})$. Each text feature $t_i$ is then combined with the image feature $v=E_v(X_{test})$ to compute the cosine similarity score:
\begin{equation}
    sim\left( t_i,v \right)=\frac{t_i \cdot v}{\Vert t_i \Vert  \Vert v \Vert}.
\end{equation}
The prediction probability $p(y_i|X_{test})$ is calculated using softmax with temperature $\tau$, as defined by:
\begin{equation}
    p(y_i|X_{test})=\frac{\exp \left( sim\left( t_i,v \right) /\tau \right)}{\sum\nolimits_{j=1}^N{\exp \left( sim\left( t_j,v \right) /\tau \right)}}.
    \label{eq:prob}
\end{equation}

\textbf{Visual Reasoning on Human Object Interaction.}
In the realm of context-dependent visual reasoning, exemplified by tasks like the Bongard-HOI task \cite{jiang2022bongard}, each test sample consists of two sets of support images and a query image, which is subsequently evaluated. These support image sets depict the presence or absence of specific human-object interaction (HOI) concepts, such as "eat orange". The model aims to identify whether the queried HOI concept exists in the query image. In this task, each concept is described as a visual relationship $c = \langle s, a, o \rangle$. In this relationship, $s$ denotes the subject (usually "human" in HOI tasks), $a$ indicates the action and $o$ represents the involved object. Each test sample $X_{test}$ encapsulates a specific concept $c = \langle s, a, o \rangle$ in one set of support images, serving as positive examples. The other set of support images acts as negative examples, illustrating a concept $\hat{c} = \langle s, \hat{a}, o \rangle$, where $\hat{a}$ is different from $a$. Notably, the task does not explicitly provide the object $o$ or the action $a$. Instead, it relies on the model's reasoning ability to predict whether the concept $c$ is present or absent in the query image. Previous studies \cite{nie2020bongard, chen2020new} have addressed the Bongard-HOI problem by training models on a diverse array of similar tasks, leveraging the Bongard-HOI training split to enable robust inference on test samples. In this context, CLIP does not require additional training data, as it already possesses comprehensive knowledge of various visual concepts. Therefore, CLIP proves to be a suitable choice for this type of visual reasoning task.

\subsection{Overview of Our Proposed NODE-Adapter}

\begin{figure*}[t]
\begin{center}
\centerline{\includegraphics[width=1\linewidth]{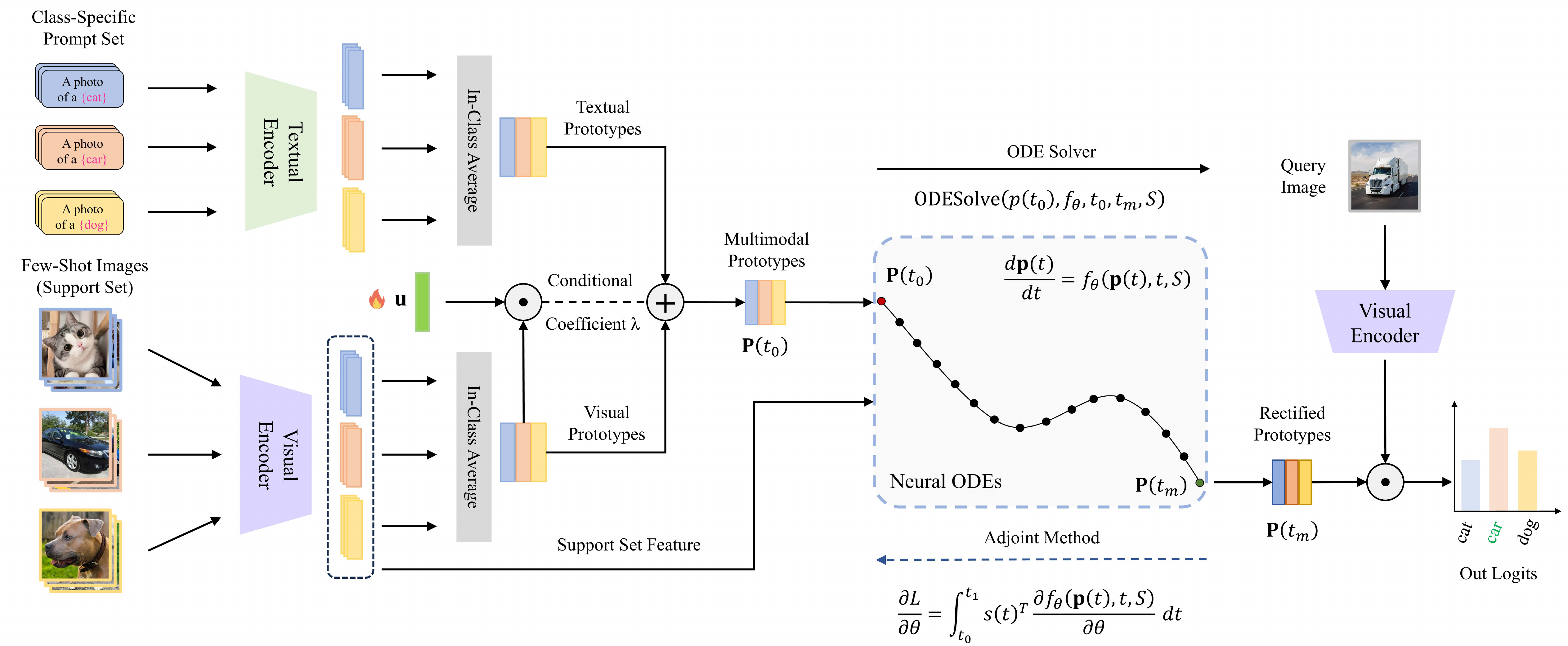}}
\caption{\textbf{An overview of our NODE-Adapter}. We first leverage the powerful aligning capability of CLIP to obtain the primitive textual and visual class prototypes. To exploit both modalities, we utilize a learnable vector $\mathbf{u}$ to conditionally combine the prototypes as the initial value for the ordinary differential equation. Then, we apply Neural ODEs to obtain the gradient and solve the initial value problem with an ODE solver as the optimal prototype to formulate the final prediction. }
\label{fig:overview}
\end{center}
\vspace{-20pt}
\end{figure*}

In figure \ref{fig:overview}, we present an overview of our proposed NODE-Adapter method. For few-shot learning with the pre-trained CLIP model and a new dataset, each of the $K$ categories contains $N$ annotated images, resulting in $N$-shot $K$-class training samples. We use $\{S_i\}_{i=1}^K$ to denote the available samples in each class. We first utilize CLIP's encoders to map images and prompts into an aligned vector space, then we  compute mean-based prototypes of the two modalities. The initial prototype $\mathbf{P}(t_0)$ is represented as a conditional combination of the two modalities.
However, the initial prototypes often exhibit bias due to the limited training data. To mitigate this issue, we design an end-to-end deep learning architecture that learns the optimal trajectory from $\mathbf{P}(t_0)$ to $\mathbf{P}(t_m)$, where $\mathbf{P}(t_0)$ represents the initial input and $\mathbf{P}(t_m)$ denotes the optimal prototype. Finally, we regard the optimal prototypes $\mathbf{P}(t_m)$ as the final prototypes, given an test image $x_t$, we first utilize Image Encoder $E_v$ to obtain the feature of $x_t$, denoted as $g_t$, then, we evaluate the class probability
that  $x_t$ belongs to class $K$ by computing
the cosine similarity between $x_t$ and $\mathbf{P}(t_m)$.
In our approach, we model the gradient flow of the gradient descent as a Neural ODEs and determine the updated weights for the neural networks using an ODE solver. Additionally, during backpropagation, we employ the adjoint sensitivity method to minimize computational costs. This framework ensures that the prototypes are refined continuously, thereby improving the overall accuracy and effectiveness of the learning process.

\vspace{-10pt}
\subsection{NODE-Adapter}
Our Proposed NODE-Adapter consist of two steps: 1). initial cross-modal prototype construction, 2) cross-modal prototype optimization with Neural ODEs. In the following, we will present each step in detail.
\subsubsection{Initial Cross-modal Prototype Construction} We now detail the two primary components involved in the process of initial cross-modal prototype construction.

\paragraph{Textual Prototype}\label{subsubsec:proto}  Following the zero-shot setting of CLIP, we first design $M$ prompts such as ``\texttt{a photo of } \{$class_j$\}" for a specific class $j$, then we place the class name in the "\{\}" to build the prompt, denoted as $\{\pi_i;class_j\}_{i=1}^M$, $\pi$ stands for the prompts. Next, we can generate the class-specific text features $T_j \triangleq \{ \mathbf{t}_{j,i} \}_{i=1}^M$ using text encoder $E_t$, denoted as, 
\begin{equation}
    \mathbf{t}_{j,i} = E_t(\{\pi_i; class_j\}) \in \mathbb{R}^{D}
    \label{eq:te},
\end{equation}
where $D$ denotes the channel number. After that, we can calculate the initial textual prototype of class $j$ by the mean $\Bar{\mathbf{t}}_j$ of text features belonging to the $j$-th class. Therefore, we calculate the textual prototype by 
\begin{equation}
    \mathbf{P}_{t}=[\Bar{\mathbf{t}}_1,\Bar{\mathbf{t}}_2,...,\Bar{\mathbf{t}}_N] \in \mathbb{R}^{N\times D}.
\end{equation}

\paragraph{Visual Prototype} 
Similarly, for a $N$-way $K$-shot task, the visual prototype for each class can be obtained by averaging the $L2$ normalized features of its $K$ images. In an $N$-class recognition problem, for each class $j$, we use CLIP's visual encoder $E_v$ to generate the visual features $\{ \mathbf{v}_{j,i} \}_{i=1}^K$. We then obtain the visual prototype by averaging these features to get $\Bar{\mathbf{v}}_j$.

\begin{equation}
       \mathbf{P}_{v}=[\Bar{\mathbf{v}}_1,\Bar{\mathbf{v}}_2,...,\Bar{\mathbf{v}}_N] \in \mathbb{R}^{N\times D}.
\end{equation}

\paragraph{Initial Cross-modal Prototype}
Inspired by \cite{xing2019adaptive}, for each class $j$, we fuse prototypes of different modalities by a convex combination to establish a new prototype $\mathbf{P}(t_0)=[\Bar{\mathbf{p}}_1,\Bar{\mathbf{p}}_2,...,\Bar{\mathbf{p}}_N] \in \mathbb{R}^{N\times D},$ and
\begin{equation}
\mathbf{p}_j = \lambda_j \cdot \bar{\mathbf{v}}_j + (1-\lambda_j) \cdot \bar{\mathbf{t}}_j,
\end{equation}
where coefficient $\lambda_j$ is conditioned on category by a learnable vector $\mathbf{u}$:
\begin{equation}
    \lambda_j=\frac{1}{1+\text{exp}(-\Bar{\mathbf{v}}_j^\top\mathbf{u})}
\end{equation}

\subsubsection{Cross-modal Prototype Optimization with Neural ODEs}
Despite the cross-modal prototypes being closer to the real class prototypes, due to limited supported data, the initial cross-modal prototypes inevitably exhibit bias. To mitigate this bias as much as possible, we use neural ODEs to optimize the initial cross-modal prototypes.  

Recent research by \cite{bu2021dynamical} has demonstrated that the iteration steps of Gradient Descent (GD) can be interpreted as an Euler discretization of an Ordinary Differential Equations (ODEs). When applying Gradient Descent to neural networks, the update rule for the neural network can be expressed:
$\mathbf{w}_{n+1} = \mathbf{w}_n - \eta\nabla L(\mathbf{w}_n)$, where \textbf{w} represents the weights of the neural network, $L$ denotes the loss function, \(\eta\) is the learning rate, and $\nabla L(\mathbf{w}_n)$ is the gradient of the loss function with respect to the weights. If we define $\frac{\mathrm{d}\mathbf{w}^{}}{\mathrm{d}t} = - \nabla L(\mathbf{w}_n)$ where $t$ is a continuous independent variable and 
$\frac{\mathrm{d}\mathbf{w}^{}}{\mathrm{d}t}$ represents the continuous gradient flow of the prototype $\mathbf{P}(t)$. We can leverage the insight that this gradient flow can be modeled as an ODEs and update it by a black-box ODE solver to facilitate a more accurate approximation of the solution trajectories. \textit{Can we design a deep learning architecture that learns the underlying dynamic system and behaves similarly to an ODEs?} Indeed, this is possible. By setting the step size of an Euler method to one, it is equivalence to a Residual Network (ResNet) : $\mathbf{X}_{n+1} = \mathbf{X}_n + f(\mathbf{X}_n, \theta)$. Taking the limit to the number of ResNet layers to infinity, which transformed discrete models into a continuous neural network, namely Neural ODEs.

This approach is well-suited to our scenario. To mitigate gradient bias, we can consider the prototype as an ODEs initial value problem and apply Neural ODEs to derive the final prototype \(\mathbf{P}(t_m)\), where the initial and final prototypes correspond to the mean-based and optimal prototypes, respectively. To utilize Neural ODEs for prototype optimization and address the gradient bias issue effectively, we need to configure the system and modify the Neural ODEs to incorporate an additional input variable S.
We assume the prototypes $\mathbf{P}(t)$ is a function dependent on time, where $t$ is a continuous time interval between $[t_0,t_m]$.
Here, $\mathbf{P}(t_0)$ represents the initial cross-modal prototype, and  $\mathbf{P}(t_m)$ denotes the optimal prototype. Consequently, we can reformulate the Neural ODEs as $\frac{\mathrm{d}\mathbf{\textbf{p}(t)}^{}}{\mathrm{d}t} = f_{\theta}(\mathbf{p}(t), t, S)$, where \(\theta\) represents some vector of learnt parameters, $S$ is support set and $f_\theta$ is a neural network parametrized by $\theta$ . Similar to a standard deep learning architecture, we perform a forward pass and backpropagation to update the optimized weights for the neural network. This framework enables us to leverage the continuous nature of Neural ODEs to refine prototypes, thus enhancing the accuracy and effectiveness of few-shot learning models by addressing the inherent bias in mean-based prototypes.

\textbf{Forward Pass:} In the forward pass, the problem is solved through integration by ODE solvers. The final prototype $\textbf{P}(t_m)$ is obtained from the initial prototype $\textbf{P}(t_0)$ by integrating the function f over the interval $[t_0,t_m]$: $\textbf{P}(t_{m}) = \textbf{P}(t_0) + \int_{t_0}^{t_m} f_{\theta}(\textbf{p}(t), t, S) \, dt= \text{ODESolve}(\mathbf{p}(t_0), f_\theta, t_0, t_m, S)$ Here, $\text{ODESolve}(\mathbf{p}(t_0),f_{\theta},t_0,t_m,S)$
denotes the use of an ODE solver to integrate $f_\theta$ from $t_0$ to $t_m$, given the initial value $\textbf{P}(t_0)$ and parameters $\theta$ and S. This method allows for accurate computation of the final prototype $\textbf{P}(t_m)$. \textbf{Backpropagation:} In the backward pass, we use another ODE solver, setting the final state $\textbf{P}(t_m)$ as the initial value. The loss function $L$ where the input is derived from the outcome of an ODE solver:  $L(\mathbf{p}(t_1)) = L\left( \mathbf{p}(t_0) + \int_{t_0}^{t_1} f_{\theta}(\mathbf{p}(t), t, S) \, dt \right) = L(\text{ODESolve}(\mathbf{p}(t_0), f_\theta, t_0, t_1, S))$. In a standard neural network, the gradient $\frac{\mathrm{d}\mathbf{p}^{}}{\mathrm{d}t}$ is computed using gradient descent, which involves summing the local gradients over all layers. However, this method can suffer from vanishing gradients and increased computational costs. In Neural ODEs, we employ the adjoint sensitivity method to address these issues. First, we need to determine the contribution of each timestep to the loss, measured by the sensitivity $\mathbf{a}(t)$, defined as $\mathbf{a}(t) = \frac{\partial L}{\partial \mathbf{p}(t)}$.  The dynamics of the sensitivity are governed by:
$ \frac{\mathrm{d}\mathbf{\mathbf{a}(t)}^{}}{\mathrm{d}t} = -\mathbf{a}(t)^T \frac{\partial f_{\theta}(\mathbf{p}(t), t, S)}{\partial \mathbf{p}(t)}. $ Updating the adjoint state is another ODEs problem, which can be solved as follows:
$\mathbf{a}(t_n) = \mathbf{a}(t_{n+1}) + \int_{t_{n+1}}^{t_n} \frac{\mathrm{d}\mathbf{a(t)}^{}}{\mathrm{d}t} $ Next, we compute the gradient of the loss with respect to the parameters $\theta$: $\frac{\partial L}{\partial \theta} = - \int_{t_{n+1}}^{t_{n}} \mathbf{a}(t)^T \frac{\partial f_{\theta}(\mathbf{p}(t), t, S)}{\partial \theta} \, dt$.

\textbf{ODE Solver.} In both the forward pass and the backward pass, an ODE solver is required. In our experiments, we selected the Runge-Kutta 4 (RK4) method due to it has a higher stability and higher order rate of convergence. The RK4 method has a convergence rate of order 4, making it an efficient choice for ODEs.

\subsection{Estimation Module Based on Gradient Flow}
\label{section3_4}

\begin{figure}[t]
\begin{center}
\centerline{\includegraphics[width=\linewidth]{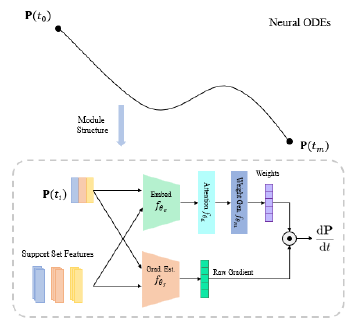}}
\caption{Structure of our Neural ODEs . With a gradient estimator and a weight generator, $f_\theta$ could adaptively capture the prototype dynamic to perform accurate rectifications.}
\label{fig:estim}
\end{center}
\end{figure}

In this section, as illustrated in Figure \ref{fig:estim}, we introduce the design of gradient estimator $f_{\theta}$ (parameterized by $\theta$). Intuitively, the prototype gradient is strongly related to the distribution of the support set $\mathcal{S}$, where each sample contains the characteristic of both the whole data domain and its class. Thus, we design a 2-stage network of $f_\theta$. The former accounts for capturing the distribution information of the dataset while the latter evaluates the contribution of samples to prototypes based on labels and then produces the corresponding weights. Finally, the gradient is estimated by a weighted sum manner. 

\paragraph{Gradient Estimator}  With the adoption of a cosine-based classifier, the prototype is expected to align with the angular center of each class. To eliminate the impact of vector norm, we first obtain the features $\mathbf{v}_i=E_{v}(\mathbf{x}_i)$ from $\mathcal{S}$, then build a class-specific representation for each sample, where prototypes $\mathbf{P}(t)$ and broadcast sample feature $\mathbf{V}_i=\mathbf{v}_i \otimes \mathds{1}_{1\times N}$ are concatenated. The representation is fed into a fully connected layer $f_{\theta_{s}}(\cdot)$ with parameters $\theta_{s}$, and the distance gradient is calculated by the difference between enhanced feature $\mathbf{v}_i$ and prototypes $\mathbf{P}(t)$:
\begin{equation} 
 	\begin{aligned}
		\mathbf{D}^{(t)}_i = f_{\theta_{s}}([\mathbf{P}(t),\mathbf{V}_i])\odot \mathbf{V}_i\ - \mathbf{P}(t),
	\end{aligned}
	\label{eq12}
\end{equation}
where $\otimes,\odot$  denote the Kronecker product for broadcasting $\mathbf{v}_i$ to all classes and the Hadamard product for element-wise multiplication, respectively. 

\paragraph{Adaptive Weight Generation}
The given label $y_i \in \mathcal{S}$ can then be leveraged to generate weight for the gradient. This is based on the rationale that prototypes should exhibit proximity to in-class features while maintaining a greater distance from others. The process involves a representation embedding layer $f_{\theta_e}(\cdot)$, a multi-head attention layer $f_{\theta_a}(\cdot)$ \cite{snell2017prototypical}, and a weight generation layer $f_{\theta_m}(\cdot)$, to obtain a robust representation by exploring the pair-wise relationship between all samples and map the them back to produce appropriate weight of $\mathbf{v}_i$:
\begin{equation}
    \begin{aligned}
            \mathbf{E}_i = f_{\theta_e}[\mathbf{P}(t),\mathbf{V}_i,\mathbf{Y}_i],\\
            \mathbf{E}_i' = f_{\theta_a}(\{E_i\}_{i=1}^{|\mathcal{S}|}),\\
            \mathbf{W}_i = f_{\theta_m}[\mathbf{E}_i'],
    \end{aligned}
\end{equation}
where the broadcast labels is $\mathbf{Y}_i=y_i \otimes \mathbf{1}_{1\times N}$. Finally, the aggregated gradient is computed by a weighted sum of gradient distance:
\begin{equation}
    \begin{aligned}
    \frac{\mathrm{d}\mathbf{P}^{}}{\mathrm{d}t}=\exp\Big(\frac{-\eta t}{T}\Big) \sum \nolimits _{i=1} ^{|\mathcal{S}|} \mathbf{W}_i \odot \mathbf{D}_i,
    \end{aligned}
\end{equation}
where  $\eta$ (which is set to 0.1 empirically) is a hyperparameter for exponential decay with time $t$ to adjust the gradient weight of different periods of optimization.

\subsection{NODE-Adapter Training and Inference}

\paragraph{NODE-Adapter Training} In the training stage, we exploit the supervised contrastive loss $\mathcal{L}_{ce}$ in a cross-entropy manner to ensure proper mapping between visual feature representation and prototypes. Given the support set $\mathcal{S}$ and prompt set, we first obtain encoded visual and textual features by CLIP's multi-modal encoders, then we can generate its visual and textual prototypes, $\mathbf{P}_v$ and $\mathbf{P}_t$, and compute their weighted sum as the initial prototype $\mathbf{P}(t_0)$. Next, we solve the ODE at time $T$ to obtain rectified prototype $\mathbf{P}(t)$. We can then evaluate the class probability that each sample $\mathbf{x_i} \in \mathcal{Q}$ belongs to class $k$ by computing the cosine similarity between its visual feature and refined prototype $\mathbf{p}_{k}(t)$, where $\mathcal{Q}$ is the query set. That is,
\begin{equation}
    P(y_i=k|\mathbf{x}_i,\mathcal{S},\mathbf{u},\theta)=\frac{\text{exp}(<E_v(\mathbf{x}_i),\mathbf{p}_{k}(t)>/\tau)}{\sum\nolimits_j \text{exp}(<E_v(\mathbf{x}_i),\mathbf{p}_{j}(t)>/\tau)},
\end{equation}
where $< \cdot >$ denotes the cosine similarity, and $\tau$ is the temperature hyper-parameter. Then, the cross-entropy loss ($\mathcal{L}_{ce}$) is computed between the predication probabilities of each input image and their corresponding class labels in the support set $\mathcal{S}$ as:
\begin{equation}
        \mathcal{L}_{ce} = \underset{\mathbf{u},\theta}{\min}  \underset{(\mathbf{x}_i,y_i) \in \mathcal{S}}{\mathbb{E}}  -\log(P(y_i|\mathbf{x}_i,\mathcal{S},\mathbf{u},\theta)),
\end{equation}
where $\theta$ is the parameters of the gradient estimator, and $\mathbf{x}_i,y_i$ denote the sample image and its corresponding class number, respectively.

\paragraph{NODE-Adapter Inference.} In the inference stage, we denote the test samples as $(\mathbf{x}_q,y_q) \in \mathcal{Q}$ for the images and their corresponding label. We compute the cosine similarity between the image feature and the rectified prototype and select the class whose prototype possesses the highest similarity:
\begin{equation}
    \hat{y_q}=\underset{y}{\arg\max} p(y|\mathbf{x}_q)=\underset{y}{\arg\max}\left<\mathbf{x}_q,\mathbf{p}_y(T)\right>
\end{equation}

\begin{table*}[htbp]
\centering
    \caption{\textbf{The detailed statistics of datasets used in experiments}. The first 11 datasets are used for few-shot learning evaluation, and the last 4 datasets are used for domain generalization.}
    \label{tab:dataset}
    \resizebox{0.87\textwidth}{!}{
    \begin{tabular}{lcccc}
    \toprule
Dataset                   &  Classes   &  Training size   &  Testing size  &  Task \\ \midrule
Caltech101~\cite{fei2004learning}  &  100  &  4,128  &  2,465 &  Object recognition \\
DTD~\cite{cimpoi2014describing} &  47  &  2,820  &  1,692  &   Texture recognition\\ 
EuroSAT~\cite{helber2019eurosat} &  10  &  13,500  &  8,100  &  Satellite image recognition \\ FGVCAircraft~\cite{maji2013fine}  &  100  &  3,334  &  3,333  &  Fine-grained aircraft recognition\\
Flowers102~\cite{nilsback2008automated}  &  102  &  4,093  &  2,463  &  Fine-grained flowers recognition \\ Food101~\cite{bossard2014food}  &  101  &  50,500 &  30,300  &  Fine-grained food recognition  \\ ImageNet~\cite{recht2019imagenet}  &  1,000  &  1.28M  &  50,000  &  Object recognition \\ OxfordPets~\cite{parkhi2012cats}  &  37   &  2,944  &  3,669  &  Fine-grained pets recognition \\ StanfordCars~\cite{krause20133d}  &  196  &  6,509  &  8,041  &  Fine-grained car recognition \\
SUN397~\cite{xiao2010sun} &  397 &  15,880  &  19,850  &  Scene recognition\\ 
UCF101~\cite{soomro2012ucf101} &  101  &  7,639  &  3,783  &  Action recognition\\
\midrule
ImageNet-V2~\cite{recht2019imagenet}  &  1,000  &  -  &  10,000  &  Robustness of collocation  \\
ImageNet-Sketch~\cite{wang2019learning}  &  1,000  &  -  & 50,889  &  Robustness of sketch domain\\
ImageNet-A~\cite{hendrycks2021natural} &  200  &  -  & 7,500  & Robustness of adversarial attack\\
ImageNet-R~\cite{hendrycks2021many} &  200  &  -  & 30,000 & Robustness of multi-domains\\
    \bottomrule
    \end{tabular}
    }
\end{table*}

\section{Experiments}
\label{sec:experiments}
In this section, we conduct experiments encompassing few-shot image classification, domain generalization and HOI visual reasoning.

\subsection{Experiment Setup} 
For the task of few-shot image classification, we adopt established methods~\cite{zhou2022learning,zhang2022tip} and implement a conventional few-shot protocol. Our approach is subjected to assessment across 11 notable image classification datasets. As shown in Table~\ref{tab:dataset}, these datasets cover a spectrum of recognition scenarios, including generic object recognition with datasets such as ImageNet~\cite{recht2019imagenet} and Caltech101~\cite{fei2004learning}; fine-grained classification involving OxfordPets~\cite{parkhi2012cats}, StandfordCars~\cite{krause20133d}, Flowers102~\cite{nilsback2008automated}, Food-101~\cite{bossard2014food}, and FGVC Aircraft~\cite{maji2013fine}; remote sensing recognition featuring EuroSAT~\cite{helber2019eurosat}; action recognition within UCF101~\cite{soomro2012ucf101}; scene recognition with SUN397~\cite{xiao2010sun}; and texture classification including DTD ~\cite{cimpoi2014describing}. This comprehensive range of datasets collectively serves as a robust benchmark for evaluating the performance of few-shot learning methods.
Regarding domain generalization, we evaluate the  model’s robustness to natural distribution shifts by training on 16-shot ImageNet~\cite{deng2009imagenet} and testing on four variants of ImageNet: ImageNet-V2~\cite{recht2019imagenet}, ImageNet-Sketch~\cite{wang2019learning}, ImageNet-A~\cite{hendrycks2021natural}, and ImageNet-R~\cite{hendrycks2021many}. These variant datasets are considered 
out-of-distribution data for ImageNet, aligning with previous researches~\cite{zhou2022conditional,shu2022tpt}. For Human object interaction visual reasoning, we follow TPT~\cite{shu2022tpt}, using Bongard-HOI \cite{jiang2022bongard} dataset.

\begin{figure*}[t]
\begin{center}
\centerline{\includegraphics[width=1\linewidth]{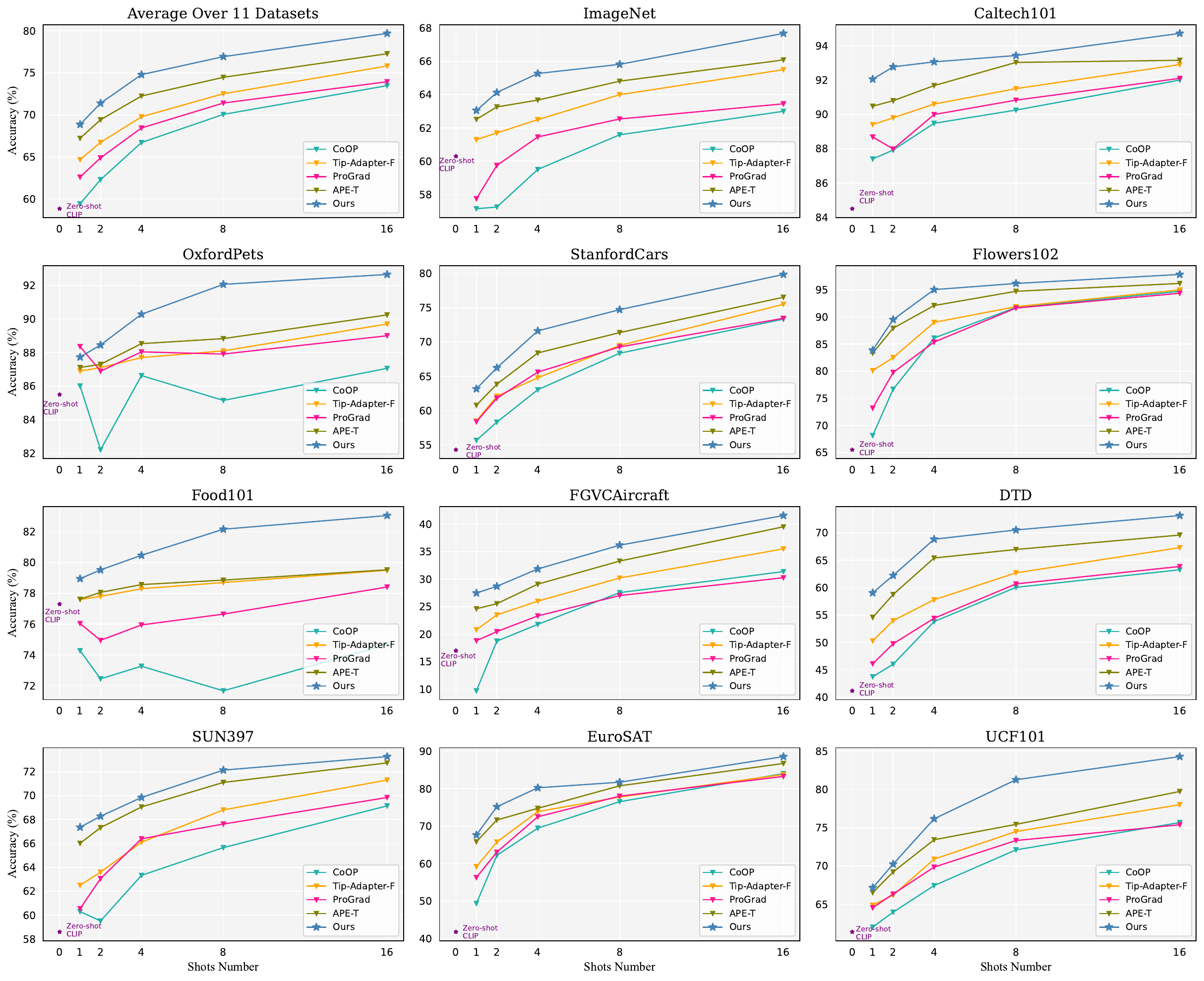}}
\caption{\textbf{Classification performance comparison on few-shot learning (on ResNet-50)}, {\em i.e.}, 1-/2-/4-/8-/16-shot, on 11 benchmark datasets. The top-left is the averaged accuracy over the 11 datasets.}
\label{fig:fewshot}
\end{center}
\end{figure*}

\textbf{Implementation Details.} 
Our approach builds upon the CLIP framework \cite{radford2021learning}, incorporating either ResNet or ViT as the image encoder and a transformer for text encoding. Notably, for few-shot classification, we use ResNet50-based CLIP for fair comparison. During training, we maintain the weights of CLIP fixed to leverage existing knowledge. Consistent with established practices, we adhere to CLIP's data pre-processing steps \cite{radford2021learning}, encompassing resizing, random cropping, and other standard operations.
In module $f_\theta$, we use a fully-connected layer with 1024-dimension output for gradient estimator $\theta_s$, a fully-connected layer with 1024-dimensional output for the embedding layer $\theta_e$, a multi-head attention module with 8 heads, where each head contains 16 units for the attention layer $\theta_a$, and a linear layer with softmax for weight generation $\theta_m$.
We set 30-epoch training for ImageNet and EuroSAT and only 20-epoch training for the other 10 datasets. The initial learning rate is set to $1\times 10^{-3}$. Parameter optimization utilizes the AdamW optimizer \cite{kingma2015adam} with a cosine annealing scheduler. Notably, our method is characterized by its parameter efficiency and lightweight nature, trained solely on a single NVIDIA RTX 3090 GPU. We manually design 20 prompts for the initial prototype calculation.

Visual reasoning on HOI differs from traditional image classification in that the accuracy of predictions in Bongard-HOI is context-dependent. Specifically, the correctness is determined by whether the example images contain the concept c or not, making it a binary evaluation. For binary labels, a simple prompting involves creating hand-craft "labels" for positive and negative examples. We use \texttt{True} for positive examples and \texttt{False} for negative ones.
We construct a hand-crafted prompt $\rho =$ "a photo that the person \{\texttt{action}\} \{\texttt{object}\}, it is \{\texttt{class}\}," where \{\texttt{action}\} represents the action, such as "drink," "play," or "sit on," \textit{etc.}; \{\texttt{object}\} refers to the object in the image, such as "football" or "chair," \textit{etc.}; and \{\texttt{class}\} is either "true" or "false." A simple example of the prompt: "a photo that person plays 
football, it is false."
\subsection{Performance Analysis} \label{sec:Performance Analysis}

\subsubsection{Few-Shot Learning} We conducted a comprehensive comparison of our proposed approach with two categories of CLIP-based adaptation methods: prompt learning methods, including CoOp and ProGrad, and adapter-style methods, specifically Tip-Adapter-F and APE-T. All these methods are built upon ResNet-50 CLIP.  Figure \ref{fig:fewshot} compares the performance of our proposed NODE-Adapter with four baseline methods on all 11 datasets. We also present the average accuracy in the top-left sub-figure of Figure \ref{fig:fewshot}. Our method surpasses all existing methods across various shot settings, except for the 1-shot OxfordPets scenario. We observe that our method outperforms other methods and obtains the highest average accuracy.
In comparison to APE-T \cite{zhang2022tip} (a fine-tuned version of APE), which is the state-of-the-art method, our method consistently exhibits substantial performance superiority across all 11 datasets. Remarkbaly, Our method with 2 shots achieve better performance than 16-shot CoOp and ProGrad on ImageNet, Caltech101 and Food101. Also, our method with 4 shots surpasses all other methods with 16 shots.
The comprehensive outcomes substantiate the effectiveness and robust performance of our proposed method.

\begin{table*}[t]
\begin{center}
    \caption{Comparison of performance on generalization (from ImageNet to ImageNet-V2/-Sketch/-A/-R) using various CLIP visual backbones.
    }
    \label{tab:DG}
    \centering
    \begin{tabular}{|l|c|c|c|c|c|c|c|c|}
        \hline
        \multirow{2}{*}{\textsc{Method}} & \multirow{2}{*}{\textsc{Visual Backbone}} & \textsc{Source} & \multicolumn{5}{c|}{\textsc{Target}} \\
        \cline{3-8}
        & & ImageNet & -V2 & -Sketch & -A & -R & Average\\
        \hline
        
        Zero-Shot CLIP \cite{radford2021learning} & \multirow{5}{*}{ResNet-50} & 58.18 & 51.34 & 33.32 & 21.65 & 56.00 & 40.58 \\
        Linear Probe CLIP \cite{radford2021learning} & & 55.87 & 45.97 & 19.07 & 12.74 & 34.86 & 28.16\\
        CoOp \cite{zhou2022learning} & & 62.95 & 55.11 & 32.74 & 22.12 & 54.96 & 41.23\\
        ProGrad \cite{zhu2022prompt}  & & 62.17  & 54.70  & 34.40  & 23.05  & 56.77  & 42.82\\
        PLOT \cite{shu2022tpt} & & 63.01  & 55.11  & 33.00  & 21.86  & 55.61  & 41.40\\
        DeFo \cite{shu2022tpt} & & 64.00  & 58.41  & 33.18  & 21.68  & 55.84  & 42.28\\
        TPT \cite{shu2022tpt}  & & 60.74  & 54.70  & 35.09  & 26.67  & 59.11  & 43.89\\
        TaskRes \cite{yu2023task} & & 64.75 & 56.47 & 35.83 & 22.80 & 60.70 & 43.90\\
        Ours & & \cellcolor[HTML]{E0FEE0}\textbf{67.68} & \cellcolor[HTML]{E0FEE0}\textbf{59.82} & \cellcolor[HTML]{E0FEE0}\textbf{38.03} & \cellcolor[HTML]{E0FEE0}\textbf{28.78} & \cellcolor[HTML]{E0FEE0}\textbf{61.82} & \cellcolor[HTML]{E0FEE0}\textbf{47.11}\\
        \hline
        
        Zero-Shot CLIP \cite{radford2021learning} & \multirow{5}{*}{ViT-B/16} & 66.73 & 60.83 & 46.15 & 47.77 & 73.96 & 57.18 \\
        Linear Probe CLIP \cite{radford2021learning} & & 65.85 & 56.26 & 34.77 & 35.68 & 58.43 & 46.29\\
        CoOp \cite{zhou2022learning} & & 71.92 & 64.18 & 46.71 & 48.41 & 74.32 & 58.41 \\
        CoCoOp \cite{zhou2022conditional}   & & 71.02  & 64.07  & 48.75  & 50.63  & 76.18  & 59.91\\
        TPT \cite{shu2022tpt}    & & 68.98  & 63.45  & 47.94  & 54.77  & 77.06  & 60.81\\
        TaskRes \cite{yu2023task} & & 73.07 & 65.30 & 49.13 & 50.37 & 77.70 & 60.63\\
        Ours & & \cellcolor[HTML]{E0FEE0}\textbf{75.85} & \cellcolor[HTML]{E0FEE0}\textbf{68.37} & \cellcolor[HTML]{E0FEE0}\textbf{53.08} & \cellcolor[HTML]{E0FEE0}\textbf{56.15} & \cellcolor[HTML]{E0FEE0}\textbf{80.26} & \cellcolor[HTML]{E0FEE0}\textbf{64.47}\\
        \hline
        Zero-Shot CLIP \cite{radford2021learning} & \multirow{5}{*}{ResNet-101} & 61.62 & 54.81 & 38.71 & 28.05& 64.38 & 46.49 \\
        Linear Probe CLIP \cite{radford2021learning} & & 59.75 & 50.05 & 26.80 & 19.44 & 47.19& 35.87\\
        CoOp \cite{zhou2022learning} & & 66.60 & 58.66 & 39.08 & 28.89 & 63.00 & 47.41 \\
        TaskRes \cite{yu2023task} & & 67.70 & 59.50 & 47.10 & 29.87 & 68.07 & 49.79\\
        Ours & & \cellcolor[HTML]{E0FEE0}\textbf{68.85} & \cellcolor[HTML]{E0FEE0}\textbf{59.93} & \cellcolor[HTML]{E0FEE0}\textbf{48.34} & \cellcolor[HTML]{E0FEE0}\textbf{30.75} & \cellcolor[HTML]{E0FEE0}\textbf{68.96} & \cellcolor[HTML]{E0FEE0}\textbf{52.00}\\
        \hline

        Zero-Shot CLIP \cite{radford2021learning} & \multirow{5}{*}{ViT-B/32} & 62.05 & 54.79 & 40.82 & 29.57 & 65.99 & 47.79 \\
        Linear Probe CLIP \cite{radford2021learning} & & 59.58 & 49.73 & 28.06 & 19.67 & 47.20 & 36.17\\
        CoOp \cite{zhou2022learning} & & 66.85 & 58.08 & 40.44 & 30.62 & 64.45 & 48.40 \\
        TaskRes \cite{yu2023task} & & 68.20 & 59.20 & 42.50 & 31.43 & 69.33 & 50.62\\
        Ours  & & \cellcolor[HTML]{E0FEE0}\textbf{70.35} & \cellcolor[HTML]{E0FEE0}\textbf{60.66} & \cellcolor[HTML]{E0FEE0}\textbf{45.08} & \cellcolor[HTML]{E0FEE0}\textbf{34.34} & \cellcolor[HTML]{E0FEE0}\textbf{71.26} & \cellcolor[HTML]{E0FEE0}\textbf{52.84}\\
        \hline
    \end{tabular}
\end{center}

\end{table*}

\subsubsection{Domain Generalization.}
Table \ref{tab:DG} summarizes the performance of our proposed method and other state-of-the-art methods. To ensure impartiality, we directly incorporate baseline results as reported in their respective original papers. It is noteworthy that, due to either the absence of reported results for specific visual backbones or compatibility issues with certain backbones, we omit comparisons with some baselines on those particular visual backbones.
Our model is trained on a 16-shot ImageNet dataset \cite{deng2009imagenet}. Subsequently, we evaluate the generalization performance of the trained models on four unseen ImageNet variants (ImageNet-V2 \cite{recht2019imagenet}, ImageNet-Sketch \cite{wang2019learning}, ImageNet-A \cite{hendrycks2021natural}, and ImageNet-R \cite{hendrycks2021many}).
As depicted in the table, our method consistently and significantly outperforms all baselines across two visual backbones (ResNet-50 and ViT-B/16). These results show the remarkable robustness of NODE-Adapter to distribution shifts.

\subsubsection{Human Object Interaction Visual Reasoning}
We include five previous methods for comparison: (1) CNN-Baseline \cite{nie2020bongard} is a simple classifier trained on the Bongard-HOI training data, where the model is trained to map an entire training sample, encompassing both the support and query images, to a binary output indicating whether the query image contains the corresponding concept; (2) Meta-Baseline \cite{chen2020new} treats each sample in the Bongard-HOI dataset as a few-shot task. The model is trained on the Bongard-HOI training data with a meta-objective designed to enable rapid adaptation to new tasks; (3) ProtoNet \cite{snell2017prototypical} acquires knowledge of a metric space that enables classification through the computation of distances from prototype representations for each class. (4) HOITrans \cite{zou2021end}, the previous best method on Bongard-HOI, is a transformer-based HOI detection model known for its state-of-the-art accuracy across various HOI detection benchmarks. It addresses Bongard-HOI by comparing the detected HOIs in the query images with those in the support images.
(5) TPT \cite{shu2022tpt} is developed upon CLIP, it can learn adaptive prompts on the fly with a single test sample.

\begin{figure*}[t!]
\centering
\includegraphics[width=1\textwidth]{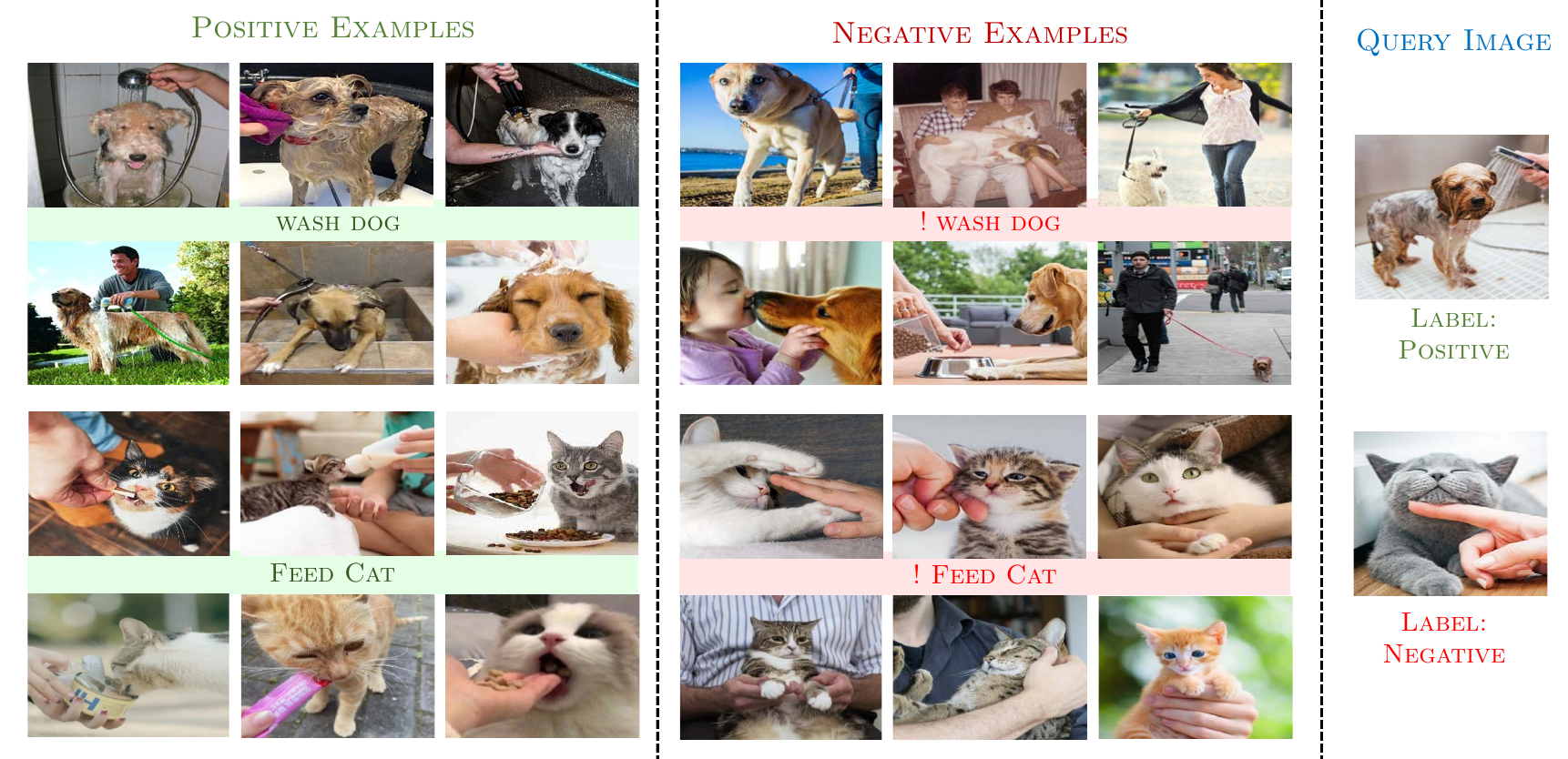}
\caption{Illustration of a Few-Shot Learning Instance from the Bongard-HOI \cite{jiang2022bongard} Benchmark: The left side features positive images illustrating the visual relationship of a person washing a dog, whereas negative examples lack this relationship. The right side presents query images, with ground-truth labels indicating whether they are positive or negative. }
\label{fig:hoi_example}
\end{figure*}

Figure \ref{fig:hoi_example} showcases several instances from the Bongard-HOI dataset \cite{jiang2022bongard}. It is important to note that each test instance actually consists of 6 positive examples, 6 negative examples, and 1 query image, which differs from the illustration shown here. As shown in Table \ref{table:hoi}, we compare the performance of the proposed NODE-Adapter approach with previous methodologies. Notably, our method significantly outperforms the traditional methods by substantial margins.  Even compared to CLIP-based TPT \cite{shu2022tpt} and BDC-Adapter \cite{zhang2023bdc}, NODE-Adapter still yields better performance. Following the experimental design detailed in Jiang et al. \cite{jiang2022bongard}, we conducted the comparison on four distinct test splits of the Bongard-HOI dataset. Notably, in the Bongard-HOI dataset, the test images are categorized into four subsets based on the presence of the HOI concept in the training data, specifically whether the action (a) or the object (o) has appeared in the training data. The results for the other baselines are directly sourced from the research paper by Jiang \textit{et al.} \cite{jiang2022bongard}, as noted. Interested readers are directed to the aforementioned paper for additional details.

\begin{figure*}[t!]
\vspace{-10pt}
\centering
\includegraphics[width=1\textwidth]{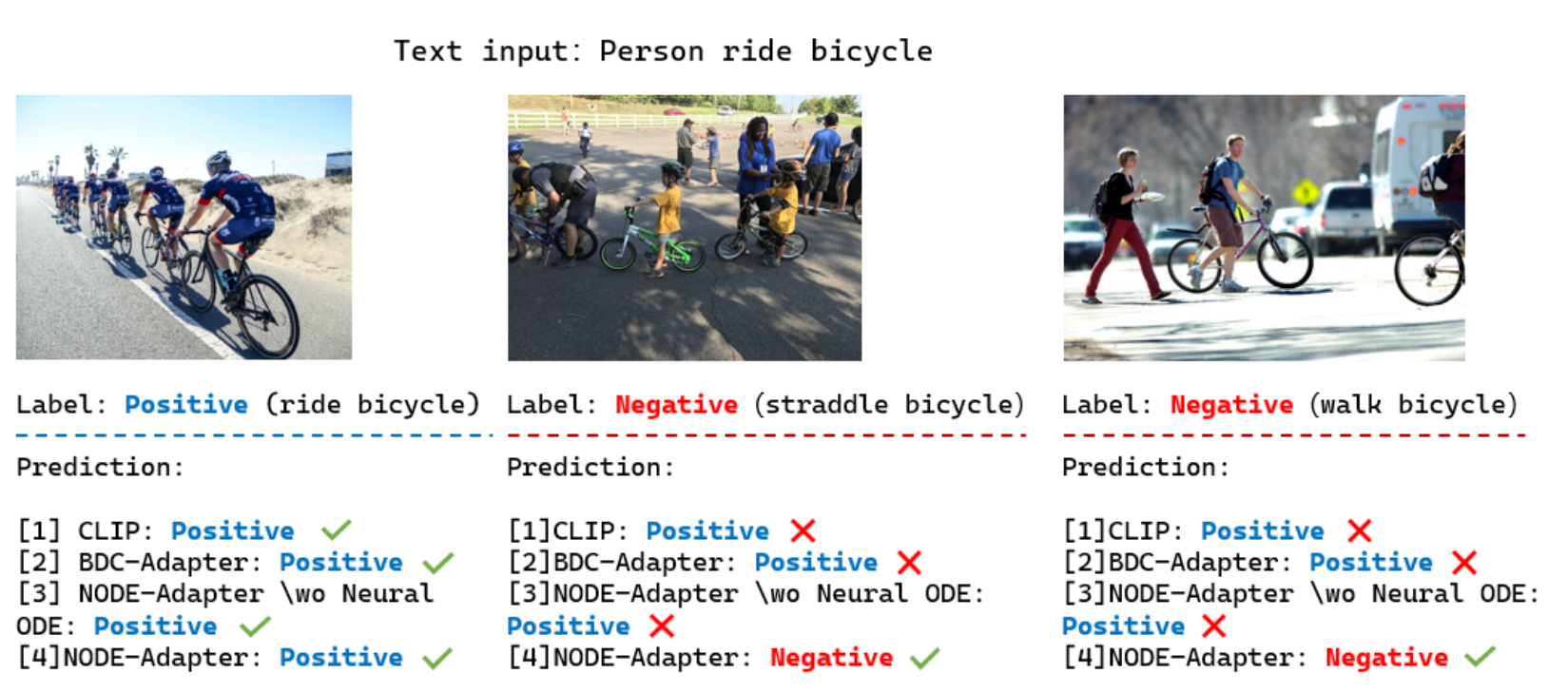}
\caption{ Cases visualization of HOI visual reasoning on the Bongard-HOI \cite{jiang2022bongard} Benchmark. Positive indicates that the 'act' words in the text input accurately reflect the interaction depicted in the image, while Negative indicates the opposite.}
\label{fig:hoi_example2}
 \vspace{-5pt}
\end{figure*}

\begin{table}[ht]
\caption{Performance comparisons of our NODE-Adapter methods and other baselines on the Bongard-HOI \cite{jiang2022bongard} dataset.}
\vspace{-0.2cm}
\label{table:hoi}
\centering
\resizebox{\linewidth}{!}{
\begin{tabular}{lccccc}
\toprule
\multirow{3}{*}{Method} & \multicolumn{5}{c}{Test Splits}          \\
\cmidrule(lr){2-6}
                        & Seen act.     & Unseen act.     & Seen act.     & Unseen act.     & \multirow{2}{*}{Avg.}    \\
                        & Seen obj.     & Seen obj.     & Unseen obj.     & Unseen obj.     &     \\
\midrule
CNN-Baseline \cite{nie2020bongard}                   & 50.03 & 49.89 & 49.77 & 50.01 & 49.92 \\
Meta-Baseline \cite{chen2020new}                   & 58.82 & 58.75 & 58.56 & 57.04 & 58.30 \\
ProtoNet  \cite{snell2017prototypical}                  & 58.90 & 58.77 & 57.11 & 58.34 & 58.28 \\
HOITrans \cite{zou2021end}                    & 59.50 & 64.38 & 63.10 & 62.87 & 62.46 \\
TPT (RN50) \cite{shu2022tpt}                    & 66.39 & 68.50 & 65.98 & 65.48 & 66.59 \\
BDC-Adapter \cite{zhang2023bdc} & 68.36 & 69.15 & 67.67 & 67.82 & 68.25 \\
Ours (RN50)    & \cellcolor[HTML]{E0FEE0}\textbf{69.45} & \cellcolor[HTML]{E0FEE0}\textbf{69.90} & \cellcolor[HTML]{E0FEE0}\textbf{68.83} & \cellcolor[HTML]{E0FEE0}\textbf{69.73} & \cellcolor[HTML]{E0FEE0}\textbf{69.48} \\
\hline
\end{tabular}
}
\end{table}

\subsection{Ablation Study}
In this section, we present an empirical analysis of our design choices and demonstrate the impact of various components of our method.
\begin{table}[t]
\centering
\caption{Effectiveness of different components in our method. TP represents the textual prototype, VP is the visual prototype, and TP + VP means the cross-modal prototype. NODE stands for Neural ODE.}
\label{table:components}
\begin{tabular}{|l|c|c|c|c|c|}
\hline
\textsc{Method} & 1    & 2  & 4  & 8 & 16 \\
\hline
Zero-shot CLIP                   & 60.33 & 60.33 & 60.33 & 60.33 & 60.33\\
TP                    & 61.65 & 61.65 & 61.65 & 61.65 & 61.65\\
VP               & 61.08 & 61.63 & 62.58 & 63.05 & 63.32 \\
TP + VP          & 62.25   & 63.02     & 64.04   & 64.71    & 65.88    \\

{TP + VP + NODE (Ours)}    & \cellcolor[HTML]{E0FEE0}\textbf{63.05} & \cellcolor[HTML]{E0FEE0}\textbf{64.13} & \cellcolor[HTML]{E0FEE0}\textbf{65.27} & \cellcolor[HTML]{E0FEE0}\textbf{65.82} & \cellcolor[HTML]{E0FEE0}\textbf{67.68} \\
\hline
\end{tabular}

\end{table}

\subsubsection{Contributions of major algorithm components}
As shown in Table \ref{table:components}, all the components employ a nearest-neighbor classification strategy. Notably, the textual prototype, derived from the average textual features of hand-crafted prompts, outperforms the zero-shot CLIP. As anticipated, the cross-modal prototype surpasses both the visual prototype and the textual prototype, showing that the cross-modal prototype is closer to the actual class prototype compared to the unimodal prototypes.
With the optimization of Neural ODEs, our method achieves approximately a 2\% performance gain with 16-shot. This result demonstrates the effectiveness of Neural ODE, pushing the cross-modal prototype even closer to the actual class prototype.

\begin{table}[t]
\caption{
Number $M$ of Prompts constructed and Integral time $T$. Experiments are conducted on 16-shot ImageNet.}
\label{table:promptnum}
\centering
\begin{tabular}{c|cccccc}
\toprule
\textbf{Value of $M$}  & 5   & 10 & \textbf{15}  & 20 & 25 & 30 \\ \midrule
\textbf{Accuracy}  & 65.25 & 66.42 & \cellcolor[HTML]{E0FEE0}\textbf{67.68} & 67.56 & 67.31 & 67.16 \\ \midrule
\textbf{Value of $T$}  & 1 & 10 & 20 & \textbf{30} & 40 & 50\\ \midrule
\textbf{Accuracy}  & 65.25 & 66.42 & 67.56 & \cellcolor[HTML]{E0FEE0}\textbf{67.68} & 67.31 & 67.16 \\ 
\bottomrule
\end{tabular}

\end{table}

\subsubsection{The number $M$ of Prompts constructed} We investigate the impact of $M$ by varying the number of prompts constructed and show the results in Table \ref{table:promptnum}. We find that our method achieves the best performance when $M = 15$. When the size continues to increase, the performance decreases since the different prompts might contain the same semantic concepts, resulting in suboptimal textual prototypes.

\subsubsection{Cases visualization of HOI visual reasoning.} As shown in figure \ref{fig:hoi_example2}. We compare our method with original CLIP \cite{radford2021learning} and BDC-Adapter \cite{zhang2023bdc}, when dealing with simpler relationships between persons and objects, all methods can correctly infer. However, when handling visually semantically similar samples, Clip and BDC-Adapter tend to make wrong prediction, whereas our NODE-Adapter exhibits excellent performance.

\subsubsection{Analysis of Integral time $T$} We conduct experiments on 16-shot ImageNet with ResNet50 CLIP. Here, we report the test accuracy of integral time from 1 to 50 in Table \ref{table:promptnum}. It can be observed that our class prototype optimization model can achieve the highest accuracy when t= 30. Hence, we set T = 30 as the default in our method.

\begin{table}[ht]
\caption{\textbf{Comparison of Accuracy (\%) and Efficiency} on 16-shot ImageNet~\cite{deng2009imagenet}. ``GFLOPs'' are calculated during training with gradient back-propagation.}
\vspace{-0.2cm}
\label{table:Accuracy}
\centering
\resizebox{\linewidth}{!}{
\begin{tabular}{lccccc}
\toprule
    Methods &Training & Epochs & GFLOPs & Param. &  Acc.   
                     
\\ \midrule
\color{gray}{\textit{Zero-shot}}\\
CLIP~\cite{radford2021learning} &- &- &- &- &60.33 
\\ \midrule
\color{gray}{\textit{Training-required}}\\
CoOp~\cite{zhou2022learning}&14 h & 200 & $>$10 & 0.01 M & 62.95  \\ 
CLIP-Adapter~\cite{gao2021clip}  & 50 min &200 & 0.004 & 0.52 M & 63.59 \\ 
Tip-Adapter-F~\cite{zhang2022tipTip-Adapter} & 5 min &20 & 0.030 & 16.3 M & 65.51  \\

\textbf{NODE-Adapter (Ours)}    & \cellcolor[HTML]{E0FEE0}\textbf{ 5 min} & \cellcolor[HTML]{E0FEE0}\textbf{30} & \cellcolor[HTML]{E0FEE0}\textbf{ 0.010} & \cellcolor[HTML]{E0FEE0}\textbf{2.08 M} & \cellcolor[HTML]{E0FEE0}\textbf{67.68} \\
\hline
\end{tabular}
}
\end{table}

\begin{table}[H]
\centering

\caption{An ablation study on various ODE solvers for 16-shot ImageNet reveals that the Fourth-Order Runge-Kutta (RK4) method outperforms Euler, Explicit Adams-Bashforth (AB), and Implicit Adams-Bashforth-Moulton (ABM) methods.
}
\label{table:ODE}
\label{table:components}
\begin{tabular}{|l|c|c|c|c|}
\hline
\textsc{Method} & Euler   & AB  & ABM  & RK4  \\
\hline
Accuracy                   & 66.93 & 67.03 & 67.42 & \cellcolor[HTML]{E0FEE0}67.68\\
\hline
\end{tabular}

\end{table}

\subsubsection{Computation Efficiency.}
We also compare the computational overhead between our method and existing approaches in Table \ref{table:Accuracy}. Our experiments utilize an NVIDIA RTX 3090 GPU, focusing on performance evaluation with the 16-shot ImageNet dataset. As shown, CoOp has the smallest number of learnable parameters but demands significant training time and GFLOPs for gradient back-propagation throughout the entire textual encoder.
Tip-Adapter-F shortens training time but significantly increases the number of learnable parameters due to the fine-tuning of the entire cache model, though it requires minimal GFLOPs for gradient computation.
Conversely, our NODE-Adapter not only achieves the highest accuracy but also exhibits superior computational efficiency: requiring \textbf{1000 times fewer GFLOPs compared to CoOp, and 8 times fewer parameters than Tip-Adapter-F.}

\subsubsection{ODE Solver}
We evaluate the performance of Euler’s method, Explicit Adams-Bashforth (AB), Implicit Adams-Bashforth-Moulton (ABM), and the fourth-order Runge-Kutta (RK4) method. Table \ref{table:ODE} presents the results for all methods on the 16-shot ImageNet dataset. Our qualitative measures align with the empirical data, demonstrating that the fourth-order Runge-Kutta method surpasses Euler, Adams-Bashforth, and Adams-Bashforth-Moulton in performance.

\section{Conclusion}
\label{sec:conclusion}
In this paper, we address the challenge of prototype-based vision-language reasoning. We propose a novel approach named \textbf{NODE-Adapter}, which leverages Neural Ordinary Differential Equations (Neural ODEs) for enhanced vision-language reasoning. Our method aims to effectively and accurately estimate class prototypes by dividing the process into two stages: cross-modal prototype construction and optimization using Neural ODEs. Specifically, we utilize Vision-Language Models (VLM) to encode hand-crafted prompts into textual features and few-shot support images into visual features. The textual and visual prototypes are derived by averaging their respective features and adaptively combining them to form the cross-modal prototype. To mitigate prototype bias, we model the prototype optimization process as an initial value problem with Neural ODEs to estimate continuous gradient flow.
Our extensive experimental evaluations cover few-shot classification, domain generalization, and visual reasoning tasks involving human-object interaction. The results convincingly demonstrate that our proposed method outperforms existing state-of-the-art approaches by a significant margin.

While our current method has demonstrated promising results, there is ample opportunity for future improvement. We plan to investigate the potential of higher order Neural ODEs, such as Second-Order Neural ODEs, for prototype optimization. Additionally, our method can be extended to address various other tasks, including image depth estimation and visual question answering. 

\section*{Acknowledgments}
CWC gratefully acknowledges funding from CCMI, University of Cambridge. CBS acknowledges support from the Philip Leverhulme Prize, the Royal Society
Wolfson Fellowship, the EPSRC advanced career fellowship EP/V029428/1, EPSRC grants EP/S026045/1 and EP/T003553/1, EP/N014588/1, EP/T017961/1, the Wellcome Innovator Awards 215733/Z/19/Z and 221633/Z/20/Z, CCMI and the Alan Turing Institute. 
AAR gratefully acknowledges funding from the Cambridge Centre for
Data-Driven Discovery and Accelerate Programme
for Scientific Discovery, made possible by a donation
from Schmidt Futures, ESPRC Digital Core Capability
Award, and CMIH, CCMI, University of Cambridge.


\bibliographystyle{IEEEtran}
\bibliography{IEEEabrv,ref}

\vfill
\end{document}